\documentclass[10pt,twocolumn,letterpaper]{article}

\usepackage[pagenumbers]{cvpr} 

\definecolor{cvprblue}{rgb}{0.21,0.49,0.74}
\usepackage[pagebackref,breaklinks,colorlinks,allcolors=cvprblue]{hyperref}
\usepackage{cuted}
\usepackage{tabularx}
\usepackage{multirow}

\def\paperID{} 
\def\confName{CVPR}
\def\confYear{2026}

\title{GazeOnce360: Fisheye-Based 360\(^\circ\) Multi-Person Gaze Estimation with Global-Local Feature Fusion}

\author{Zhuojiang Cai \qquad Zhenghui Sun \qquad Feng Lu\textsuperscript{\rm} 
\thanks{Corresponding Author. This work was supported by the National Natural Science Foundation of China (NSFC) under Grant 62372019.}\\
State Key Laboratory of VR Technology and Systems, School of CSE, Beihang University\\
{\tt\small \{caizhuojiang, sunzhenghui, lufeng\}@buaa.edu.cn}
}

\begin{document}
\maketitle

\begin{strip}
    \centering
    \vspace{-4em}
    \centering
    \includegraphics[width=\textwidth]{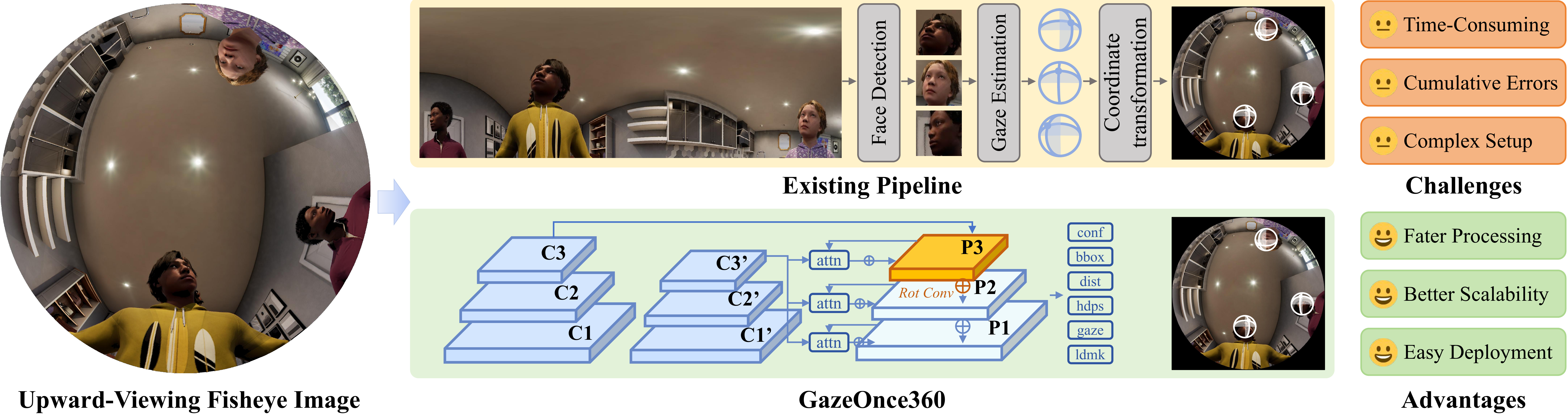}
    \captionof{figure}{\textbf{Comparison between the existing method~\cite{caiGam360SensingGaze2025} and the proposed GazeOnce360.} The existing multi-step pipeline (top) is compared against the proposed end-to-end GazeOnce360 model (bottom), highlighting the transition from a multi-step pipeline to a more efficient and scalable end-to-end solution with improved robustness and simplicity.}
    \label{fig:teaser}
\end{strip}

\begin{abstract}
We present \textbf{GazeOnce360}, a novel end-to-end model for multi-person gaze estimation from a single tabletop-mounted upward-facing fisheye camera. Unlike conventional approaches that rely on forward-facing cameras in constrained viewpoints, we address the underexplored setting of estimating the 3D gaze direction of multiple people distributed across a 360° scene from an upward fisheye perspective. To support research in this setting, we introduce \textbf{MPSGaze360}, a large-scale synthetic dataset rendered using Unreal Engine, featuring diverse multi-person configurations with accurate 3D gaze and eye landmark annotations.
Our model tackles the severe distortion and perspective variation inherent in fisheye imagery by incorporating rotational convolutions and eye landmark supervision. To better capture fine-grained eye features crucial for gaze estimation, we propose a dual-resolution architecture that fuses global low-resolution context with high-resolution local eye regions. Experimental results demonstrate the effectiveness of each component in our model. This work highlights the feasibility and potential of fisheye-based 360° gaze estimation in practical multi-person scenarios. Project page: https://caizhuojiang.github.io/GazeOnce360/.
\end{abstract}    
\section{Introduction}
\label{sec:intro}

Understanding human gaze is essential for applications in human–computer interaction~\cite{aliasghariImplementingGazeControl2020, belardinelliGazeBasedIntentionEstimation2024}, collaboration~\cite{pfeufferMultiuserGazebasedInteraction2021}, virtual reality~\cite{plopskiEyeExtendedReality2023, baoExploring3DInteraction2023}, and behavioral analysis~\cite{harezlakApplicationEyeTracking2018, kasneciIntroductionEyeTracking2024, keUsingEyetrackingEducation2024}.
Over the past decade, significant progress has been made in \emph{single-person} gaze estimation, driven by large-scale datasets such as EYEDIAP~\cite{funesmoraEYEDIAPDatabaseDevelopment2014}, MPIIGaze~\cite{zhangMPIIGazeRealWorldDataset2017}, and ETH-XGaze~\cite{zhangETHXGazeLargeScale2020}.
These datasets have enabled a wide range of appearance-based gaze regression methods~\cite{balujaNonIntrusiveGazeTracking1993, tanAppearancebasedEyeGaze2002, luHeadPosefreeAppearancebased2012, luLearningGazeBiases2014, zhangAppearanceBasedGazeEstimation2015, wangGeneralizingEyeTracking2019, chengAppearanceBasedGazeEstimation2024, ghoshAutomaticGazeAnalysis2024}, achieving robust 3D gaze prediction under both controlled and in-the-wild conditions.
While this line of work has largely matured for isolated individuals, real-world scenarios often involve multiple people interacting within the same scene, motivating the problem of \emph{multi-person} gaze estimation.
Prior efforts in multi-person gaze estimation have primarily relied on forward-facing camera setups, where individuals appear in relatively constrained viewpoints~\cite{zhangGazeOnceRealTimeMultiPerson2022}.
Such setups capture only a limited field of view and often require multiple synchronized devices to ensure full coverage, which makes deployment cumbersome and restricts applicability in everyday environments.

In this work, we explore a more compact and flexible sensing configuration: a single upward-facing fisheye camera placed on a tabletop, capturing the entire 360\textdegree{} surrounding scene from below.
This panoramic viewpoint is particularly suitable for interactive applications in everyday spaces, such as collaborative workspaces, reception counters, and service robots.
Existing attempts to perform multi-person gaze estimation under this configuration typically rely on complex multi-step pipelines~\cite{caiGam360SensingGaze2025}, which are time-consuming, difficult to deploy, and prone to cumulative errors.
For example, projecting fisheye images into panoramic representations can occasionally split faces across boundaries, leading to missed detections.
This motivates an end-to-end approach that directly predicts multi-person gaze from fisheye input while addressing the specific challenges of this setup.
These challenges include severe geometric distortion caused by the fisheye lens, redundant computation from high-resolution background regions, and the lack of publicly available multi-person upward-facing fisheye gaze datasets.

To address these challenges, we propose GazeOnce360, a novel dual-resolution end-to-end model for 360\textdegree{} multi-person gaze estimation in fisheye-from-below settings.
To support supervised training, we also construct MPSGaze360, a large-scale synthetic dataset rendered using Unreal Engine, containing diverse multi-person arrangements, realistic illumination, and precise annotations for face bounding boxes, 3D gaze, head pose, and eye landmarks.
This dataset enables training in scenarios that would be prohibitively difficult to annotate in the real world.

GazeOnce360 addresses the core challenges of fisheye multi-person gaze estimation in three ways.
First, we employ rotational convolutions~\cite{weiRotationalConvolutionRethinking2023} to improve geometric consistency under fisheye distortion and extreme viewing angles.
Second, we introduce \textit{multi-task supervision} leveraging face and eye landmarks to guide the network toward spatially meaningful ocular representations.
Third, we design a \textit{dual-resolution architecture} that fuses global low-resolution context with local high-resolution eye regions, enabling efficient yet fine-grained gaze reasoning across multiple subjects.

Extensive experiments validate the effectiveness of each component in GazeOnce360, showing that rotational convolution, multi-task supervision, and dual-resolution fusion each contribute substantially to performance.
Qualitative results on real fisheye images demonstrate the model's generalization capability and highlight its potential for practical multi-person gaze estimation in unconstrained real-world environments.

Our contributions are summarized as follows:
\begin{itemize}
\item We introduce MPSGaze360, the first large-scale synthetic dataset designed for multi-person 360\textdegree{} gaze estimation under a tabletop fisheye configuration, with high-fidelity 3D gaze, head, and eye landmark annotations.
\item We propose GazeOnce360, a dual-resolution fisheye-based multi-person gaze estimation model that effectively handles the challenges of fisheye imagery.
\item We validate our approach through extensive quantitative and qualitative experiments, demonstrating the effectiveness of each component in our model and its generalization to real fisheye images.
\end{itemize}
\section{Related Work}

\subsection{Gaze Estimation Datasets}

Large annotated datasets have played a central role in the advancement of appearance-based gaze estimation. Representative benchmarks range from controlled to in-the-wild collections, with EYEDIAP~\cite{funesmoraEYEDIAPDatabaseDevelopment2014} and MPIIGaze~\cite{zhangMPIIGazeRealWorldDataset2017} providing early controlled and real-world samples. RT-GENE~\cite{fischerRTGENERealTimeEye2018} and Gaze360~\cite{kellnhoferGaze360PhysicallyUnconstrained2019} further expand variability in illumination and head pose, while ETH-XGaze~\cite{zhangETHXGazeLargeScale2020} offers high-resolution coverage of extreme head and gaze angles captured by multi-camera rigs. These resources have significantly contributed to robust single-person gaze estimation in unconstrained environments. 

Synthetic datasets have also been explored to supplement real-world gaze data.
MPSGaze~\cite{zhangGazeOnceRealTimeMultiPerson2022} employs an image-replacement and compositing pipeline to create multi-person scenes from existing datasets, allowing supervised training without manual annotation.
GazeGene~\cite{baoGazeGeneLargescaleSynthetic2025} uses Unreal Engine to render photorealistic single-person gaze images and has demonstrated that models trained on such synthetic data can generalize to real-world scenarios. 
However, all existing datasets are captured using perspective cameras, and thus differ substantially from the visual distributions in upward-facing fisheye settings.

\subsection{Gaze Estimation Methods}

Appearance-based gaze estimation predominantly relies on deep convolutional architectures to regress gaze directions from eye crops or full-face images of a single person.
Since the introduction of convolutional method by Zhang \textit{et al.}~\cite{zhangAppearanceBasedGazeEstimation2015}, many works have improved gaze estimation networks along various directions~\cite{zhangItsWrittenAll2017, chengAppearanceBasedGazeEstimation2018, chenAppearanceBasedGazeEstimation2019, ogusuLPMLearnablePooling2019, wangGeneralizingEyeTracking2019, chengCoarsetoFineAdaptiveNetwork2020, baoAdaptiveFeatureFusion2021, baoFeatureGazeGeneralizable2024a}, including more robust feature extraction, attention mechanisms, and feature fusion strategies.
More recently, transformer-based architectures have also been applied to single-person gaze estimation~\cite{chengGazeEstimationUsing2022, qinUniGazeUniversalGaze2025}, demonstrating competitive performance in unconstrained settings.

End-to-end multi-person gaze estimation methods have also been explored, which integrate face detection and gaze regression within a single network, avoiding repeated per-face cropping and redundant feature extraction.
For example, GazeOnce~\cite{zhangGazeOnceRealTimeMultiPerson2022} employs an anchor-based detection head combined with multi-task supervision to achieve real-time multi-person gaze estimation.

While these appearance-driven approaches achieve strong performance for single-person scenarios and can scale to multiple subjects, they are designed for perspective cameras and do not directly address the challenges of fisheye imagery.
GAM360~\cite{caiGam360SensingGaze2025} attempts to perform multi-person gaze estimation from upward-facing fisheye images using a multi-stage pipeline, but a more compact, and easily deployable end-to-end solution has not yet been explored.

\subsection{Fisheye Imaging and Perception}

Fisheye cameras introduce severe radial distortions, which have been addressed through both geometric modeling and learning-based adaptation.
Classical projection models such as the equidistant model and the Kannala-Brandt (KB) polynomial~\cite{kannalaGenericCameraCalibration2004} provide accurate mappings between image pixels and viewing directions, forming the basis for downstream visual perception tasks.
In our work, we adopt the equidistant model for fisheye projection, and note that our synthetic dataset can be conveniently reprojected using calibrated KB parameters to facilitate training distortion-adapted models.

Fisheye cameras are widely used in many computer vision applications, including surround-view perception for autonomous driving~\cite{kumarSurroundviewFisheyeCamera2023} and a variety of indoor tasks where the camera is mounted either on the ceiling in a top-down configuration or on robots in upward-facing settings~\cite{liWeaklySupervisedMultiPersonAction2020, chiangEfficientPedestrianDetection2021, pullingGeometryInformedDistanceCandidate2024}.
Prior works in these domains typically process fisheye images in one of two ways.
A common strategy is to reproject the fisheye image into multiple perspective views before feeding them into standard CNNs~\cite{liWeaklySupervisedMultiPersonAction2020, chiangEfficientPedestrianDetection2021}, which restores translation equivariance but often introduces discontinuities and distortions near view boundaries.
Another direction proposes network architectures inherently compatible with omnidirectional distortions, including rotation-invariant or rotation-equivariant convolution operators~\cite{cohenSphericalCNNs2019, zhuDeformableConvNetsV22018, weiRotationalConvolutionRethinking2023}, which directly operate on fisheye images without requiring unfolding or reprojection.
Our work follows the latter direction by incorporating rotational convolution~\cite{weiRotationalConvolutionRethinking2023} into our model, and we demonstrate its effectiveness for the challenging task of multi-person gaze estimation in upward-facing fisheye imagery.
\section{Problem Formulation}
\label{sec:problem}

We address the problem of multi-person 3D gaze estimation from a single upward-facing fisheye camera. The camera is placed on a horizontal surface (e.g., a tabletop) and oriented vertically upwards, capturing a 360\textdegree{} panoramic view of its surroundings. The goal is to estimate the 3D gaze direction vector $\mathbf{g}_i \in \mathbb{R}^3$ for every visible person $i$ in the input fisheye image $\mathbf{I} \in \mathbb{R}^{H \times W \times 3}$, where all gaze vectors are defined in the camera coordinate system.

Formally, our model learns a mapping function $\mathcal{F}$ that takes a fisheye image $\mathbf{I}$ as input and predicts the set of 3D gaze directions for all $N$ detected individuals:
\begin{equation}
    \{\mathbf{g}_i\}_{i=1}^{N} = \mathcal{F}(\mathbf{I}; \Theta),
\end{equation}
where $\Theta$ denotes the learnable parameters of the model.

The fisheye image $\mathbf{I}$ is modeled under an \textit{equidistant projection}, where the radial distance $r$ from the image center is linearly proportional to the incident angle $\theta$ of the incoming light ray:
\begin{equation}
    r = f \cdot \theta,
\end{equation}
with $f$ denoting the focal length. In this work, we assume a 180\textdegree{} field of view (FOV), which allows the camera to capture the entire surrounding scene while introducing severe geometric distortions near the periphery.

Our objective is to design an efficient and deployable fisheye-based multi-person gaze estimation framework that can accurately predict gaze directions under strong fisheye distortions and wide-angle spatial configurations.
\section{Synthetic Dataset: MPSGaze360}
\label{sec:dataset}

\begin{figure}[t]
\centering
\includegraphics[width=\columnwidth]{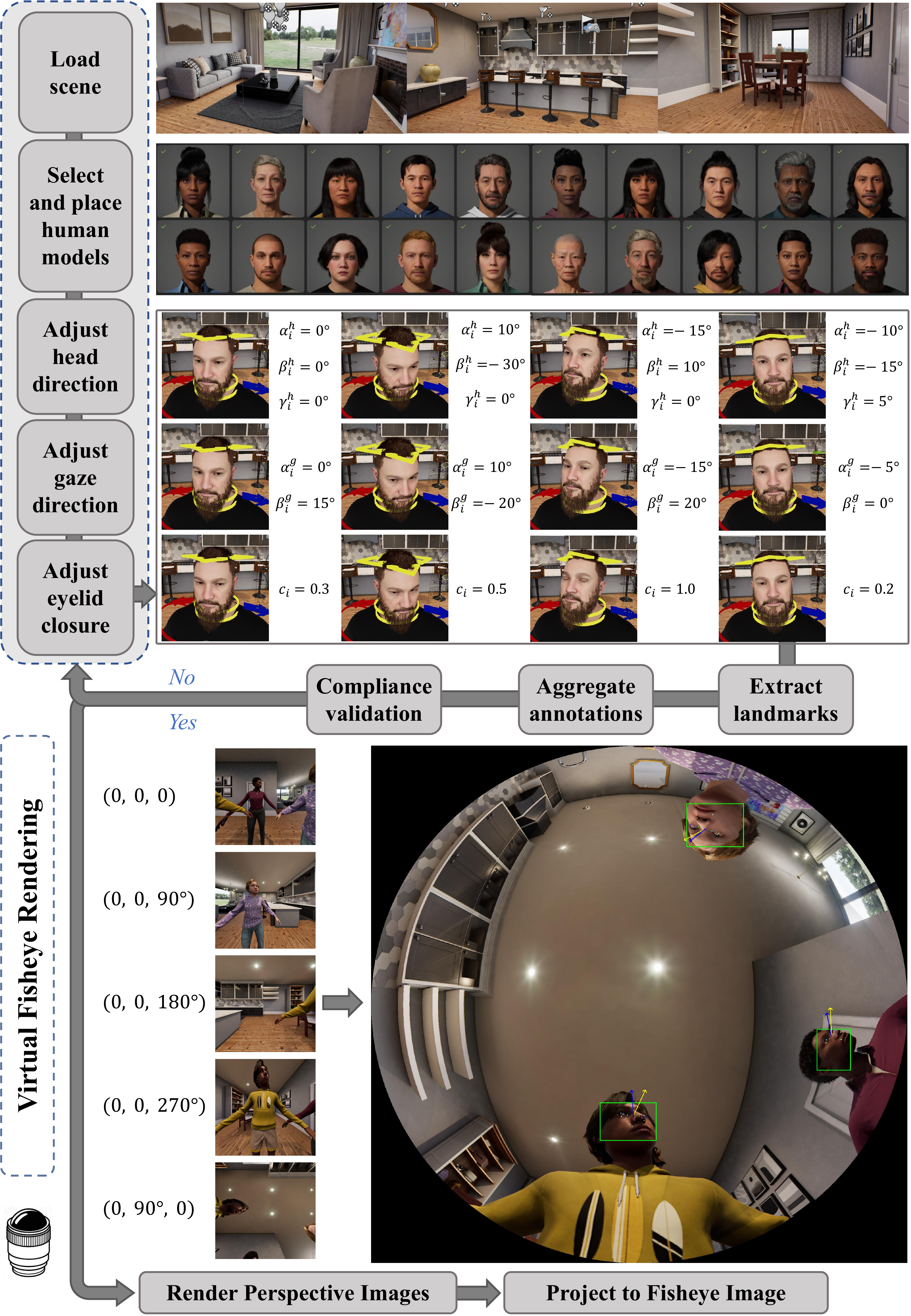}
\caption{
\textbf{Overview of the MPSGaze360 data generation pipeline.}
We first load a virtual indoor environment and populate it with multiple MetaHuman characters of diverse appearances.
For each subject, we randomly sample head orientation $(\alpha_i^h, \beta_i^h, \gamma_i^h)$, gaze direction $(\alpha_i^g, \beta_i^g)$, and eyelid closure $c_i$. The process also includes extraction of 2D landmarks and gaze vectors, annotation aggregation, and compliance validation.
Each sample is rendered as five orthogonal perspective views and subsequently projected into a single 180\textdegree{} equidistant fisheye image.}
\label{fig:dataset_pipeline}
\vspace{-2mm}
\end{figure}

To support supervised training in the proposed fisheye-based multi-person gaze estimation framework, 
we construct a synthetic dataset named \textbf{MPSGaze360} (Multi-Person Synthetic Gaze 360°) using Unreal Engine 5 (UE5) with MetaHuman to generate photorealistic environments and human renderings. 
An overview of the complete data generation pipeline is illustrated in Fig.~\ref{fig:dataset_pipeline}, 
which visualizes the major steps including scene initialization, virtual human placement, gaze and head pose sampling, and fisheye projection.
Our virtual setup replicates a realistic tabletop scene, where a fisheye camera is placed at table height, facing upwards, to capture a 360\textdegree{} panoramic view of multiple surrounding people. 
Individuals are positioned at varying distances and orientations, each gazing toward different directions to simulate natural social interactions. 
Within this controlled environment, we render high-quality images and automatically record precise ground-truth annotations.

\subsection{Generation Pipeline}

To ensure both diversity and realism, we randomly sample key scene parameters within physically reasonable ranges.
This includes the number of people, their spatial arrangement around the camera, and their body, head, and eye orientations, as well as eyelid closure.
Head orientations are uniformly sampled over yaw angles of $[-60^\circ, 60^\circ]$, pitch angles of $[-35^\circ, 35^\circ]$, and roll angles of $[-5^\circ, 5^\circ]$.
The 3D gaze directions are sampled in the local head coordinate system with yaw and pitch angles drawn from $[-30^\circ, 30^\circ]$.
We further randomize lighting conditions and person-related appearance attributes, including skin tone, age, clothing, and hairstyle, to enhance domain diversity.

Since Unreal Engine does not provide a native fisheye camera, all images are rendered as five orthogonal perspective views and subsequently projected into a single 180\textdegree{} equidistant fisheye image.

\subsubsection{Ground-Truth Annotations}

For each rendered frame, we provide comprehensive ground-truth annotations to facilitate both gaze estimation and auxiliary supervision tasks. These include: (1) 3D gaze direction vectors for every visible individual, expressed in the camera coordinate system; (2) 2D face and eye landmarks projected onto the image plane; (3) face bounding boxes; (4) 3D head poses represented by rotation and translation relative to the camera; and (5) distance from the camera. All annotations are automatically extracted from the scene, ensuring pixel-level accuracy and consistency.

\subsection{Dataset Statistics}

\begin{figure}[h]
\centering
\includegraphics[width=\columnwidth]{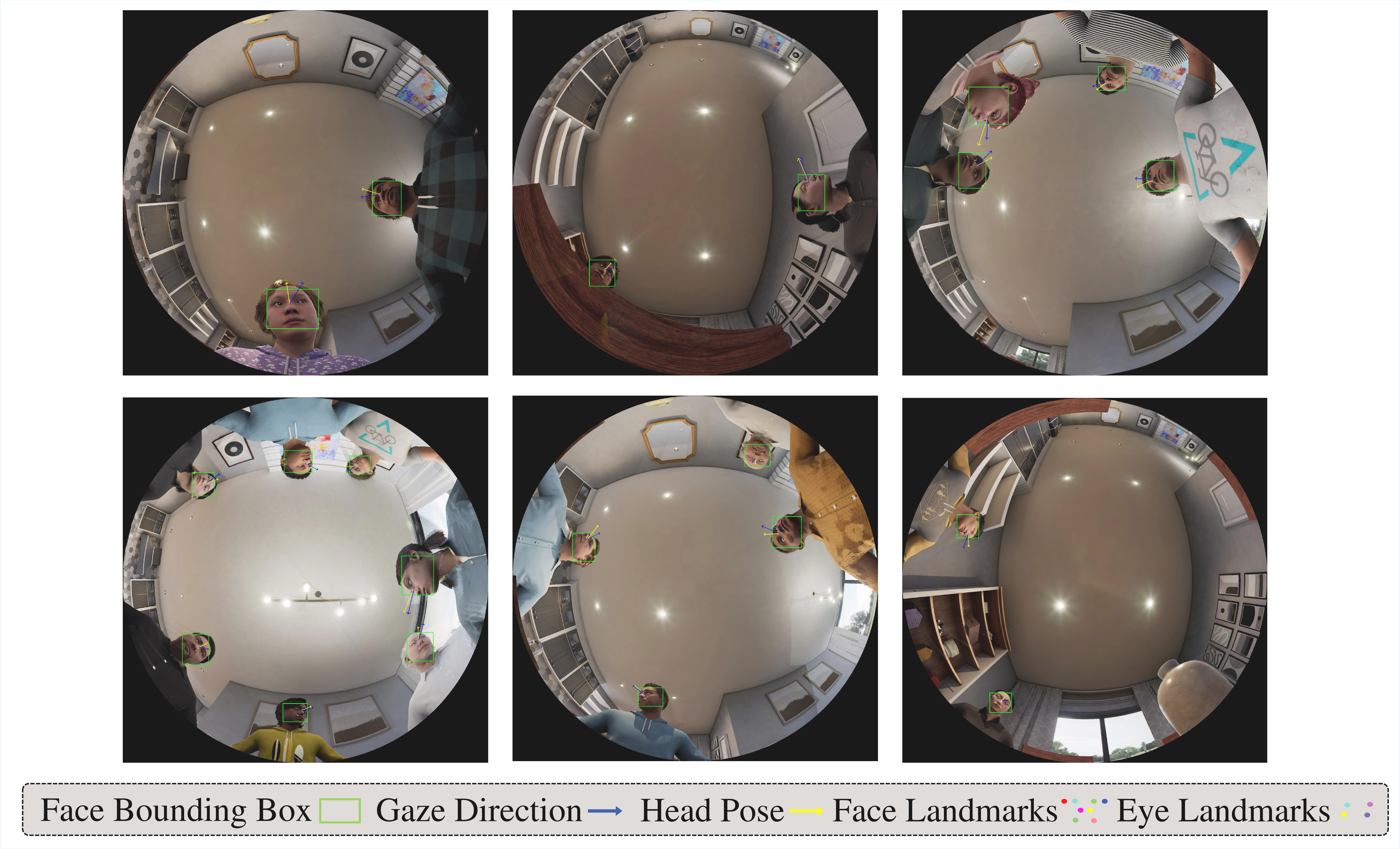}
\caption{
\textbf{Visualization of the proposed MPSGaze360 dataset.}
These examples demonstrate the realism, diversity, and annotation accuracy of the synthetic dataset.
}
\label{fig:dataset_samples}
\vspace{-3mm}
\end{figure}

MPSGaze360 comprises 23,496 synthesized fisheye images, each containing 1-7 faces. In total, the dataset includes 69 distinct human models, covering a wide range of genders, ethnicities, and age groups. Figure~\ref{fig:dataset_samples} presents representative examples with rendered scenes and their corresponding annotations. The diversity in camera configurations, head poses, lighting, and subject appearances makes MPSGaze360 a challenging and comprehensive benchmark for fisheye-based 360\textdegree{} multi-person gaze estimation.

\section{GazeOnce360}
\label{sec:method}

\begin{figure*}[t]
\centering
\includegraphics[width=0.95\textwidth]{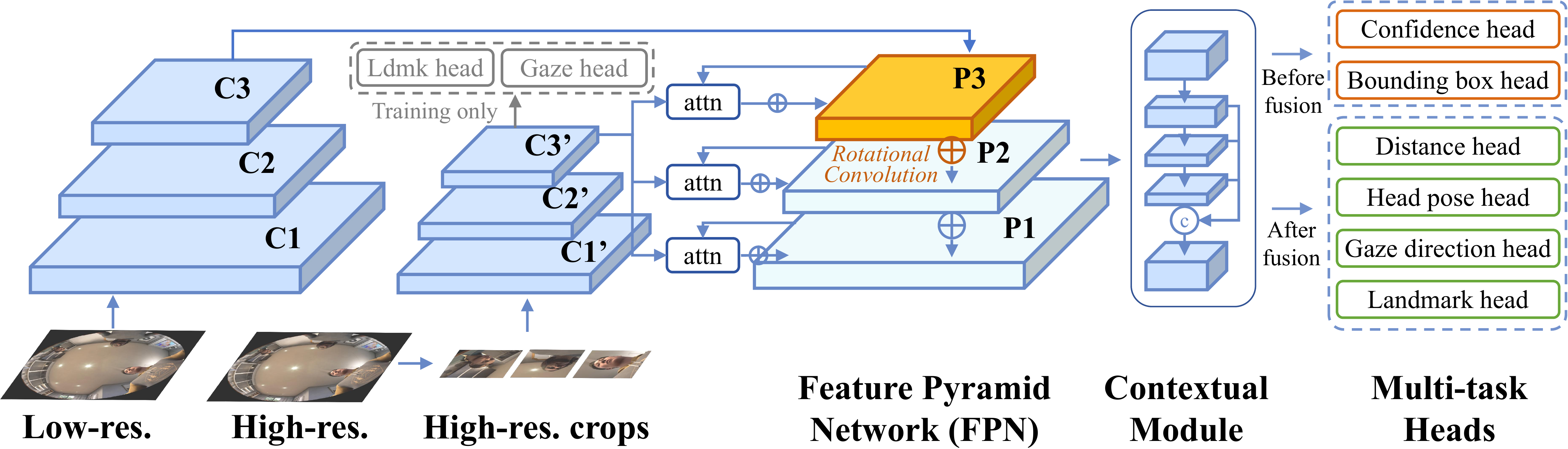}
\caption{
\textbf{Overall architecture of GazeOnce360.}
GazeOnce360 is an anchor-based CNN with rotational convolutions.
The low-resolution fisheye image is processed by the \emph{global branch} to capture large-scale spatial context and to detect face bounding boxes.
For each detected region, the \emph{local branch} extracts high-resolution facial features to capture fine-grained gaze cues.
A cross-attention module fuses global and local representations, followed by multi-task heads that predict landmarks and gaze directions.
During training, an additional supervision signal is applied to the local branch to regress per-face gaze for improved ocular feature learning.
}
\label{fig:framework}
\end{figure*}

In this section, we introduce the GazeOnce360 network. As illustrated in Figure~\ref{fig:framework}, the network combines a multi-person gaze estimation model with rotational convolution and multi-task supervision, along with an efficient dual-resolution architecture that fuses high- and low-resolution features. Specifically, a low-resolution global branch extracts contextual information and predicts face bounding boxes, while a high-resolution branch processes cropped faces. The features from both branches are then fused via cross-attention to produce the gaze predictions. The model is trained end-to-end in a single stage: during training, the local face branch uses ground-truth bounding boxes for cropping, whereas at test time it relies on predicted boxes.

\subsection{Preliminaries: Rotational Convolution}
\label{sec:rotconv_prelim}

Rotational convolution is a previously proposed technique aimed at enhancing a network’s rotation invariance, which is particularly important for fisheye imagery captured from extreme viewpoints such as top-down or upward-facing configurations~\cite{weiRotationalConvolutionRethinking2023}.
Unlike standard convolution kernels that are inherently translation invariant, rotational convolution augments this by replicating a kernel into four rotated versions corresponding to orthogonal orientations.
The feature responses from these rotated kernels are then aggregated through a weighted averaging process, enabling the network to better accommodate large angular variations introduced by fisheye distortions.

Following Wei \emph{et~al.}~\cite{weiRotationalConvolutionRethinking2023}, we integrate rotational convolution into the top level of the Feature Pyramid Network (FPN), allowing high-level features to benefit from improved rotation-invariant representations.
As demonstrated in our ablation experiments, incorporating rotational convolution provides clear performance gains for multi-person gaze estimation under fisheye projections.

\subsection{Dual-Resolution Feature Fusion}
\label{sec:dual_resolution_fusion}

In multi-person gaze estimation, high-frequency facial cues are far more informative than high-resolution background details, yet global scene structure remains essential for disambiguating multi-person interactions. To jointly leverage these complementary sources of information, we adopt a dual-resolution feature fusion scheme composed of two parallel branches. The \textit{global branch} processes a low-resolution fisheye image to capture overall spatial layout, crowd arrangement, and coarse person-to-camera geometry. The \textit{local branch} operates on high-resolution facial crops to extract fine-grained ocular and periocular features that are critical for accurate gaze prediction. Both branches follow standard CNN backbones; the key contribution lies in the cross-resolution fusion mechanism described below.

\paragraph{Feature Extraction.}
Let the high-resolution branch output feature map $\mathbf{F}_h \in \mathbb{R}^{N \times H_h \times W_h \times C_h}$, where $N$ is the number of detected faces. 
To reduce spatial redundancy while preserving semantic information, global average pooling is applied:
\begin{equation}
\bar{\mathbf{F}}_h = \frac{1}{H_h W_h}\sum_{i=1}^{H_h}\sum_{j=1}^{W_h} \mathbf{F}_h(i,j),
\label{eq:avgpool}
\end{equation}
yielding the compact facial descriptor $\bar{\mathbf{F}}_h \in \mathbb{R}^{N\times C_h}$.  
For the global branch, the feature map at the $s$-th FPN scale is $\mathbf{F}_{l,s} \in \mathbb{R}^{H_{l,s}\times W_{l,s}\times C_l}$. 
After adding positional encoding $\mathbf{P}_{h,s}$, it is flattened into a sequence representation as
\begin{equation}
\tilde{\mathbf{F}}_{l,s} = \operatorname{flatten}\!\big(\mathbf{F}_{l,s} + \mathbf{P}_{h,s}\big),
\label{eq:flatten_pos}
\end{equation}
where $N_{l,s}=H_{l,s}\times W_{l,s}$ denotes the total number of spatial positions.

\paragraph{Cross-Attention Fusion.}
To integrate the two sources, a cross-attention mechanism is employed.  
The low-resolution features act as queries $\mathbf{Q}=\mathrm{MLP}_Q(\tilde{\mathbf{F}}_{l,s})$, and the high-resolution facial features serve as keys $\mathbf{K}=\mathrm{MLP}_K(\bar{\mathbf{F}}_h)$ and values $\mathbf{V}=\mathrm{MLP}_V(\bar{\mathbf{F}}_h)$.  
The attention operation is defined as
\begin{equation}
\operatorname{Attention}(\mathbf{Q},\mathbf{K},\mathbf{V}) =
\operatorname{softmax}\!\left(\frac{\mathbf{Q}\mathbf{K}^\top}{\sqrt{d_k}}\right)\mathbf{V},
\label{eq:attention}
\end{equation}
where $d_k$ is the feature dimension of the key vectors.  
This operation models correlations between each spatial position in the global feature map and the compact facial descriptors, enabling adaptive emphasis on gaze-relevant regions.

\paragraph{Feature Integration.}
The fused representation at scale $s$ is obtained through residual aggregation:
\begin{equation}
\mathbf{F}_{\mathrm{fuse},s} =
\mathbf{F}_{l,s} + \mathbf{M}\odot
\operatorname{Attention}(\mathbf{Q},\mathbf{K},\mathbf{V}),
\label{eq:fuse}
\end{equation}
where $\mathbf{M}$ is a spatial mask restricting attention to the corresponding facial regions on the low-resolution feature map, and $\odot$ denotes element-wise multiplication.  
This residual form preserves global spatial context while injecting high-resolution semantic cues from facial features.

\paragraph{Spatial Mask.} During cross-attention between local face features and global image features, we apply a spatial mask $\mathbf{M}$ for each face (one mask per face) corresponding to its region on the global feature map. This mechanism helps the model focus on the most discriminative regions.

\paragraph{Multi-Scale Aggregation.}
The above process is performed across multiple FPN scales ($s\in\{1,2,3\}$), producing fused features $\{\mathbf{F}_{\mathrm{fuse},s}\}$. Features at different scales are further processed by a contextual module and then fed into a multi-task head that predicts face confidence, bounding boxes, head pose, and gaze direction. 

\subsection{Multi-Task Supervision}

The output of this network, \( y \), can be expressed as the set \( \{y_c, y_b, y_d, y_h, y_g, y_{fl}, y_{el}\} \). Among these outputs, \( y_c \) (confidence), \( y_b \) (bounding box), and \( y_g \) (gaze direction) are the key outputs for multi-person gaze estimation tasks. \( y_d \) (head distance) and \( y_h \) (head pose) are additional information required in gaze interaction systems, and both are predicted simultaneously as part of the network's output. Landmarks \( y_{fl} \) (facial landmarks) and \( y_{el} \) (eye landmarks) serve as additional supervision to improve the model's accuracy and generalization.

Notably, the training data for facial and eye landmark annotations come from the precise coordinates of the MetaHuman human model used in the data synthesis process. The eye landmarks include the centers of both eyes and the pupils, which are difficult to obtain from real-world data. Some studies attempt to infer eye landmark annotations through gaze and pupil center labels, but this method often introduces additional annotation errors. Thanks to the use of synthetic data generation, this study uses the precise eye landmarks to train the deep learning model, improving gaze estimation accuracy through multi-task supervision.

\subsection{Loss Functions}

The network is trained using a multi-task joint optimization strategy, and the overall loss function is composed of the weighted losses from each task head:
{\small
\begin{equation}
\mathcal{L} = \lambda_1 \mathcal{L}_c + \lambda_2 \mathcal{L}_b + \lambda_3 \mathcal{L}_d + \lambda_4 \mathcal{L}_h + \lambda_5 \mathcal{L}_g + \lambda_6 \mathcal{L}_{fl} + \lambda_7 \mathcal{L}_{el}
\end{equation}
}

Where \( \mathcal{L}_c \) is the classification loss, which uses a balanced cross-entropy loss for positive and negative samples. The other losses \( \{\mathcal{L}_b, \mathcal{L}_d, \mathcal{L}_h, \mathcal{L}_g, \mathcal{L}_{fl}, \mathcal{L}_{el}\} \) all use Smooth L1 loss on positive samples. The weights \( \lambda_1, \lambda_2, \dots, \lambda_7 \) control the contribution of each task to the total loss.

\section{Experiment}

In this section, we comprehensively evaluate the proposed \textbf{GazeOnce360} network for multi-person gaze estimation in fisheye images. 
The experiments aim to validate three key aspects of our approach: 
(1) the effectiveness of rotational convolution for handling omnidirectional distortions, 
(2) the benefit of multi-task supervision with facial and ocular landmarks, and 
(3) the dual-resolution architecture that balances global context and local detail.

We first conduct controlled experiments on the standard evaluation split to validate the proposed components, and then evaluate the generalization ability of the model under cross-identity and cross-scene settings.

\subsection{Implementation Details}
\label{sec:implementation}

All experiments were conducted on a subset of the MPSGaze360 dataset, which contains 5,673 training and 804 testing fisheye images rendered under equidistant projection at a resolution of $1024 \times 1024$.
Both the global and local branches adopt ResNet-50 backbones, sharing the same architecture but maintaining separate parameters.
The network was trained on three NVIDIA RTX 2080 Ti GPUs for 150 epochs with a batch size of 9. We used the Adam optimizer with an initial learning rate of $10^{-3}$, which was reduced to $10^{-4}$ and $10^{-5}$ at the 30th and 100th epochs, respectively.
The entire model was trained end-to-end, and all results were evaluated on the same test split.

\subsection{Evaluation Metrics}

We adopt four standard metrics to assess both localization and directional accuracy:
(1) \textbf{Detection performance:} bounding box precision and recall;
(2) \textbf{Localization accuracy:} head distance error (cm);
(3) \textbf{Orientation accuracy:} head pose and gaze direction angular error (degrees);
(4) \textbf{Runtime:} average inference frame rate (FPS).
Among these, the gaze direction angular error \textbf{(Gaze Error)} serves as the principal indicator. To improve the comparability across different experimental settings, we also report an \textbf{Adjusted Gaze Error} metric in some experiments. 
This metric computes the gaze error only over the intersection of successfully detected faces shared by the compared methods, thereby decoupling gaze estimation accuracy from potential differences in detection results.

\subsection{Model Validation Experiments}

We first conduct a series of controlled experiments on the standard evaluation split of the dataset to analyze the contributions of the proposed architectural components.

\subsubsection{RotConv and Multi-Task Supervision}
\label{sec:ablation}

We first analyze the effectiveness of the rotational convolution (RotConv) and the supervision strategy. 
Table~\ref{tab:ablation_rot_mt} presents quantitative results across different configurations.

\begin{table}[htbp]
\centering
\small
\setlength{\tabcolsep}{3pt}
\renewcommand{\arraystretch}{1.1}
\caption{Ablation results for RotConv and landmarks supervision.}
\label{tab:ablation_rot_mt}
\resizebox{\columnwidth}{!}{
\begin{tabular}{lccccc}
\hline
Setting & Prc.$\uparrow$ & Rec.$\uparrow$ & Dist (cm)$\downarrow$ & Head pose (°)$\downarrow$ & Gaze (°)$\downarrow$ \\
\hline
w/o RotConv, w/o ldmks     & 0.9836 & 0.9927 & 3.486 & 5.010 & 12.14 \\
RotConv only            & 0.9923 & \underline{0.9934} & 3.422 & 4.150 & 11.14 \\
RotConv + face ldmks     & 0.9888 & 0.9932 & 3.447 & 3.769 & 9.782 \\
RotConv + eye ldmks      & \underline{0.9936} & \textbf{0.9940} & \textbf{3.387} & \underline{3.448} & \textbf{8.890} \\
RotConv + face \& eye ldmks & \textbf{0.9981} & 0.9933 & \underline{3.390} & \textbf{3.411} & \underline{8.945} \\
\hline
\end{tabular}}
\end{table}

The results show that introducing RotConv alone already reduces both head pose and gaze errors compared to the baseline. 
Further incorporating landmark supervision, especially from eye landmarks, provides additional gains by refining ocular feature learning. 
The best overall accuracy (gaze error $= 8.89^\circ$) is achieved when combining RotConv with eye landmark supervision.

\begin{figure}[t]
\centering
\includegraphics[width=\columnwidth]{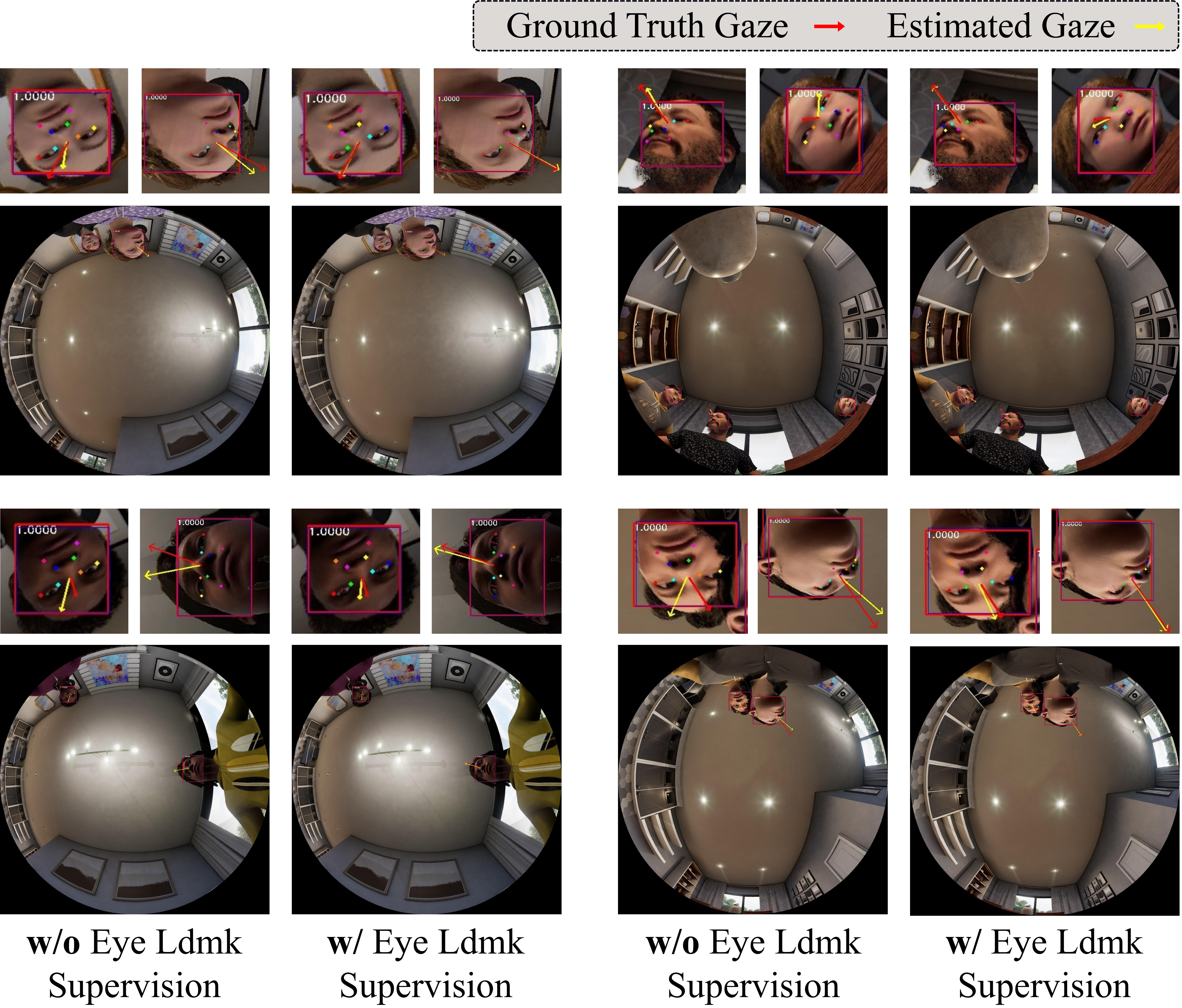}
\caption{
\textbf{Qualitative comparison of gaze prediction with and without eye landmark supervision.}
Each column shows predicted gaze directions (yellow) and ground truth (red).
Without eye landmark supervision (left), predictions deviate under extreme head poses, 
while adding landmark supervision (right) captures fine-grained ocular cues for more accurate gaze directions.
}
\label{fig:eye_supervision}
\vspace{-3mm}
\end{figure}

As shown in Fig.~\ref{fig:eye_supervision}, the addition of eye landmark supervision substantially improves the alignment between predicted and true gaze directions across diverse scenes. 
This validates that fine-grained eye geometry cues enhance the learning of orientation-sensitive representations.





\subsubsection{Dual-Resolution Feature Fusion}
\label{sec:dual_resolution}

We further compare the proposed dual-resolution architecture against single-resolution baselines. 
Table~\ref{tab:res_comparison} summarizes the trade-off between accuracy and efficiency.

\begin{table}[htbp]
\centering
\small
\setlength{\tabcolsep}{3pt}
\renewcommand{\arraystretch}{1.1}
\caption{Comparison between single- and dual-res. networks.}
\label{tab:res_comparison}
\resizebox{\columnwidth}{!}{
\begin{tabular}{lcccccc}
\hline
Setting & Prc.$\uparrow$ & Rec.$\uparrow$ & Dist (cm)$\downarrow$ & Head pose (°)$\downarrow$ & Gaze (°)$\downarrow$ & FPS$\uparrow$ \\
\hline
Single (512)     & 0.9960 & 0.9915 & 3.389 & 3.659 & 16.50 & 20.49 \\
Single (1024)    & 0.9981 & 0.9933 & 3.390 & 3.411 & 8.945 & 13.30 \\
Dual (512,1024)  & 0.9986 & 0.9933 & 3.652 & 3.235 & 8.968 & 16.23 \\
\hline
\end{tabular}}
\end{table}

The dual-resolution network attains nearly identical gaze accuracy to the high-resolution model (8.968° vs. 8.945°) 
while maintaining a 22\% higher inference speed. 
This confirms that combining low-resolution global context with high-resolution local refinement 
achieves an optimal balance between efficiency and precision, 
making the method more scalable for real-time multi-person scenarios.

\subsection{Generalization Experiments}

To evaluate the generalization ability of the proposed model, we conduct additional experiments on the cross-scene + cross-identity evaluation dataset.

\subsubsection{Cross-Identity and Cross-Scene Evaluation}

The cross-identity setting (D1) evaluates the model on identities that do not appear in the training set, while the cross-scene + cross-identity setting (D2) additionally ensures that both the scene and identities are unseen during training.

\begin{table}[htbp]
\centering
\small
\caption{Cross-scene and cross-identity evaluation.}
\label{tab:cross_scene_identity}
\resizebox{\columnwidth}{!}{
\begin{tabular}{lccc}
\hline
Test Dataset & Precision $\uparrow$ & Recall $\uparrow$ & Gaze Error ($^\circ$) $\downarrow$ \\
\hline
Non-cross Scene \& Identity (Baseline) & 0.9981 & 0.9933 & 8.945 \\
Cross-Identity (D1)                    & 0.9997 & 0.9956 & 9.446 \\
Cross-Scene + Cross-Identity (D2)      & 0.9992 & 0.9903 & 10.39 \\
\hline
\end{tabular}}
\end{table}

Compared with the standard evaluation split, the gaze error increases moderately under the D2 setting, indicating that the model maintains stable performance even under significant domain shifts.

\subsubsection{Comparison with GAM360}

We further compare the proposed method with GAM360~\cite{caiGam360SensingGaze2025} under the challenging D2 evaluation setting.

\begin{table}[htbp]
\centering
\small
\caption{Quantitative comparison with GAM360.}
\label{tab:gam360_comp}
\resizebox{\columnwidth}{!}{
\begin{tabular}{lccc}
\hline
Method & Gaze Error ($^\circ$) $\downarrow$ & Adjusted Gaze Error ($^\circ$) $\downarrow$ & FPS $\uparrow$  \\ 
\hline
GAM360 & 18.96 & 18.76 & 4.23  \\
GazeOnce360 & \textbf{10.39}  & \textbf{9.99} & \textbf{16.23}  \\ 
\hline
\end{tabular}}
\end{table}

GAM360~\cite{caiGam360SensingGaze2025} is a multi-stage approach that first detects faces in fisheye images, projects each face to a frontal perspective, and then estimates gaze for each face separately. 
As shown in Table~\ref{tab:gam360_comp}, GazeOnce360 outperforms GAM360 in both accuracy and efficiency, reducing gaze error while achieving a substantially higher inference speed.

\begin{figure*}[htbp]
  \centering
  \includegraphics[width=0.95\linewidth]{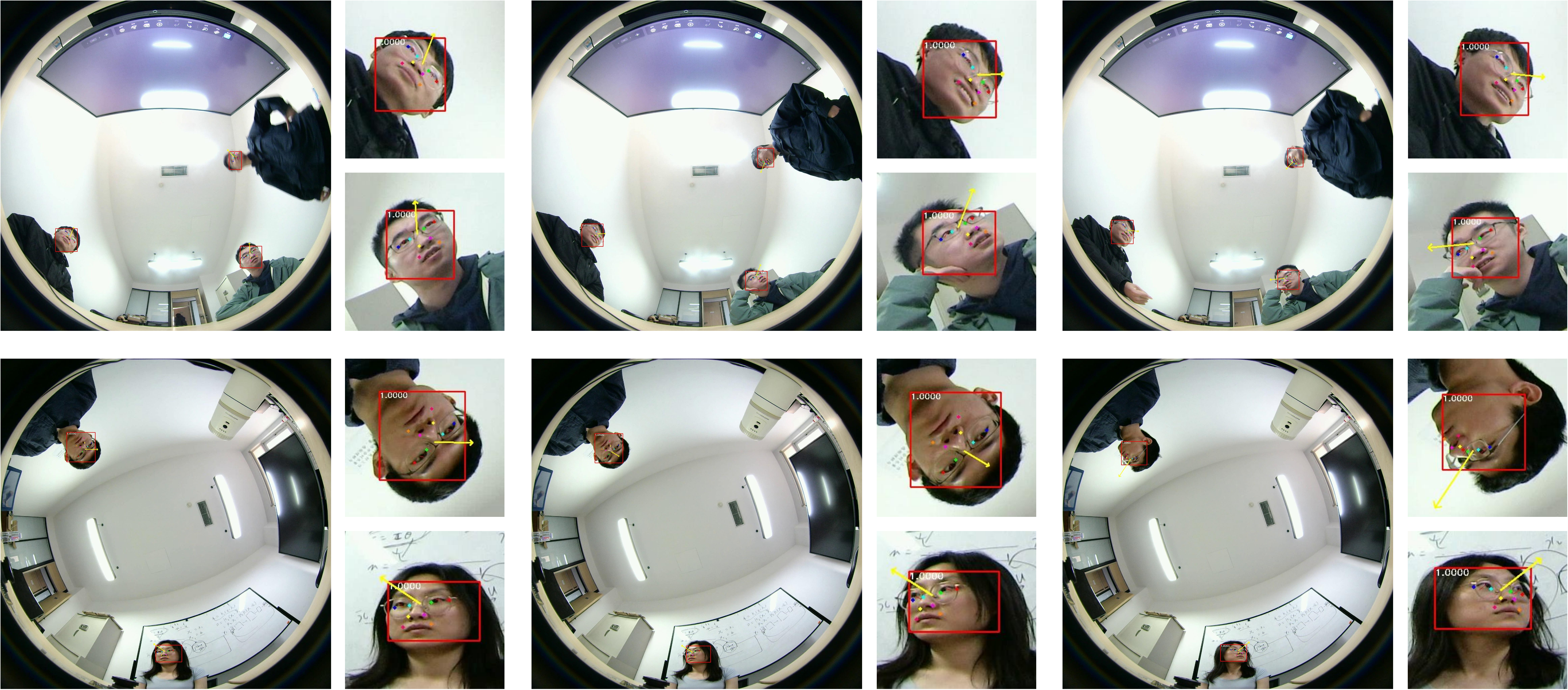}
  \caption{\textbf{Qualitative results on real fisheye images.} 
  Visualization of gaze estimation results obtained by our model on real-world fisheye camera captures. 
  Despite being trained purely on synthetic data, our method generalizes well to real scenes, accurately capturing multi-person gaze directions across diverse individuals and head poses.}
  \label{fig:real_test}
\end{figure*}

\subsubsection{RotConv vs Deformable Convolution}

To investigate whether RotConv can be replaced by other flexible convolution operators, we compare it with Deformable Convolution (DCN) under identical settings.

\begin{table}[h]
\centering
\small
\caption{Quantitative comparison with Deformable Convolution.}
\label{tab:dcn}
\resizebox{\columnwidth}{!}{
\begin{tabular}{lcccc}
\hline
Module & Precision $\uparrow$ & Recall $\uparrow$ & Gaze Error ($^\circ$) $\downarrow$ & Adjusted Gaze Error ($^\circ$) $\downarrow$ \\
\hline
RotConv & \textbf{0.9992} & \textbf{0.9903} & \textbf{10.39} & \textbf{10.37} \\
DCN     & 0.9981 & 0.9898 & 11.05 & 11.02 \\
\hline
\end{tabular}}
\end{table}

As shown in Table~\ref{tab:dcn}, RotConv consistently achieves lower gaze error, suggesting that explicitly modeling rotational distortions in fisheye images is more effective than relying solely on spatially adaptive kernels.

\subsection{Discussion}

Our extensive experiments demonstrate the effectiveness of the proposed model in terms of architectural design and training strategies, validate its generalization ability, and highlight its advantages over multi-stage approaches. 

In particular, the rotational convolution adapts effectively to the characteristics of fisheye images for gaze estimation, outperforming both standard and deformable convolutions. The dual-resolution design is motivated by the challenges of fisheye imagery: the larger field of view causes faces to occupy a smaller fraction of the image compared to standard cameras, making a dedicated high-resolution branch more important. Beyond RotConv, the architecture is fully compatible with conventional cameras. 

Our multi-task training strategy, particularly supervision on eye landmarks, further improves performance, benefiting from the precise annotations available in the synthetic dataset. Interestingly, adding both eye and face landmark supervision does not outperform using eye landmarks alone. This may be because face landmarks primarily encode coarse head geometry and can introduce noise or conflicting supervision beyond the eye landmarks.

In the main experiments, we set all loss weights to 1 to avoid any task-specific weighting heuristics. In the supplementary material, we show that adjusting the gaze loss weight can further improve the model's performance.

\section{Conclusion}
\label{sec:conclusion}

In this paper, we presented GazeOnce360, a novel end-to-end network for multi-person gaze estimation from an upward-facing fisheye camera.
Moving beyond the constrained viewpoints of forward-facing cameras, our method tackles the challenging 360° gaze estimation setting through rotational convolutions for distortion robustness, eye landmark supervision for spatial guidance, and a dual-resolution architecture that fuses global context with local eye features.
To support this research direction, we introduced MPSGaze360, a large-scale synthetic dataset featuring diverse multi-person configurations and accurate gaze annotations.
Extensive experiments demonstrate the effectiveness of each component in our proposed method and its generalization to real fisheye images.

\textbf{Limitations and Future Work.}
Our method’s performance decreases for subjects with extreme head poses or large distances from the camera.
The current framework also relies primarily on synthetic data for training; although high-quality synthetic data contributes to a certain level of generalization to real-world scenarios, there remains room for improvement.
Future work will focus on enhancing cross-domain robustness, as well as improving performance under severe occlusions, extreme poses, and challenging illumination conditions.
We believe that GazeOnce360 establishes a solid foundation for practical multi-person gaze estimation in unconstrained real-world environments.

{
    \small
    \bibliographystyle{ieeenat_fullname}
    \bibliography{main}

\begin{thebibliography}{41}
\providecommand{\natexlab}[1]{#1}
\providecommand{\url}[1]{\texttt{#1}}
\expandafter\ifx\csname urlstyle\endcsname\relax
  \providecommand{\doi}[1]{doi: #1}\else
  \providecommand{\doi}{doi: \begingroup \urlstyle{rm}\Url}\fi

\bibitem[Aliasghari et~al.(2020)Aliasghari, Taheri, Meghdari, and Maghsoodi]{aliasghariImplementingGazeControl2020}
Pourya Aliasghari, Alireza Taheri, Ali Meghdari, and Ehsan Maghsoodi.
\newblock Implementing a gaze control system on a social robot in multi-person interactions.
\newblock \emph{SN Applied Sciences}, 2\penalty0 (6):\penalty0 1135, 2020.

\bibitem[Baluja and Pomerleau(1993)]{balujaNonIntrusiveGazeTracking1993}
Shumeet Baluja and Dean Pomerleau.
\newblock Non-{{Intrusive Gaze Tracking Using Artificial Neural Networks}}.
\newblock In \emph{Advances in {{Neural Information Processing Systems}}}. Morgan-Kaufmann, 1993.

\bibitem[Bao and Lu(2024)]{baoFeatureGazeGeneralizable2024a}
Yiwei Bao and Feng Lu.
\newblock From {{Feature}} to {{Gaze}}: {{A Generalizable Replacement}} of {{Linear Layer}} for {{Gaze Estimation}}.
\newblock In \emph{Proceedings of the {{IEEE}}/{{CVF Conference}} on {{Computer Vision}} and {{Pattern Recognition}}}, pages 1409--1418, 2024.

\bibitem[Bao et~al.(2021)Bao, Cheng, Liu, and Lu]{baoAdaptiveFeatureFusion2021}
Yiwei Bao, Yihua Cheng, Yunfei Liu, and Feng Lu.
\newblock Adaptive {{Feature Fusion Network}} for {{Gaze Tracking}} in {{Mobile Tablets}}, 2021.

\bibitem[Bao et~al.(2023)Bao, Wang, Wang, and Lu]{baoExploring3DInteraction2023}
Yiwei Bao, Jiaxi Wang, Zhimin Wang, and Feng Lu.
\newblock Exploring {{3D Interaction}} with {{Gaze Guidance}} in {{Augmented Reality}}.
\newblock In \emph{2023 {{IEEE Conference Virtual Reality}} and {{3D User Interfaces}} ({{VR}})}, pages 22--32, Shanghai, China, 2023. IEEE.

\bibitem[Bao et~al.(2025)Bao, Wang, and Lu]{baoGazeGeneLargescaleSynthetic2025}
Yiwei Bao, Zhiming Wang, and Feng Lu.
\newblock {{GazeGene}}: {{Large-scale Synthetic Gaze Dataset}} with {{3D Eyeball Annotations}}.
\newblock In \emph{2025 {{IEEE}}/{{CVF Conference}} on {{Computer Vision}} and {{Pattern Recognition}} ({{CVPR}})}, pages 18749--18759, 2025.

\bibitem[Belardinelli(2024)]{belardinelliGazeBasedIntentionEstimation2024}
Anna Belardinelli.
\newblock Gaze-{{Based Intention Estimation}}: {{Principles}}, {{Methodologies}}, and {{Applications}} in {{HRI}}.
\newblock \emph{J. Hum.-Robot Interact.}, 13\penalty0 (3):\penalty0 31:1--31:30, 2024.

\bibitem[Cai et~al.(2025)Cai, Wang, Niu, and Lu]{caiGam360SensingGaze2025}
Zhuojiang Cai, Haofei Wang, Yuhao Niu, and Feng Lu.
\newblock Gam360: Sensing gaze activities of multi-persons in 360 degrees.
\newblock \emph{CCF Transactions on Pervasive Computing and Interaction}, 2025.

\bibitem[Chen and Shi(2019)]{chenAppearanceBasedGazeEstimation2019}
Zhaokang Chen and Bertram~E. Shi.
\newblock Appearance-{{Based Gaze Estimation Using Dilated-Convolutions}}.
\newblock In \emph{Computer {{Vision}} -- {{ACCV}} 2018}, pages 309--324, Cham, 2019. Springer International Publishing.

\bibitem[Cheng and Lu(2022)]{chengGazeEstimationUsing2022}
Yihua Cheng and Feng Lu.
\newblock Gaze {{Estimation}} using {{Transformer}}.
\newblock In \emph{2022 26th {{International Conference}} on {{Pattern Recognition}} ({{ICPR}})}, pages 3341--3347, Montreal, QC, Canada, 2022. IEEE.

\bibitem[Cheng et~al.(2018)Cheng, Lu, and Zhang]{chengAppearanceBasedGazeEstimation2018}
Yihua Cheng, Feng Lu, and Xucong Zhang.
\newblock Appearance-{{Based Gaze Estimation}} via {{Evaluation-Guided Asymmetric Regression}}.
\newblock In \emph{Computer {{Vision}} -- {{ECCV}} 2018}, pages 105--121, Cham, 2018. Springer International Publishing.

\bibitem[Cheng et~al.(2020)Cheng, Huang, Wang, Qian, and Lu]{chengCoarsetoFineAdaptiveNetwork2020}
Yihua Cheng, Shiyao Huang, Fei Wang, Chen Qian, and Feng Lu.
\newblock A {{Coarse-to-Fine Adaptive Network}} for {{Appearance-Based Gaze Estimation}}.
\newblock \emph{Proceedings of the AAAI Conference on Artificial Intelligence}, 34\penalty0 (07):\penalty0 10623--10630, 2020.

\bibitem[Cheng et~al.(2024)Cheng, Wang, Bao, and Lu]{chengAppearanceBasedGazeEstimation2024}
Yihua Cheng, Haofei Wang, Yiwei Bao, and Feng Lu.
\newblock Appearance-{{Based Gaze Estimation With Deep Learning}}: {{A Review}} and {{Benchmark}}.
\newblock \emph{IEEE Transactions on Pattern Analysis and Machine Intelligence}, 46\penalty0 (12):\penalty0 7509--7528, 2024.

\bibitem[Chiang et~al.(2021)Chiang, Wang, and Chen]{chiangEfficientPedestrianDetection2021}
Sheng-Ho Chiang, Tsaipei Wang, and Yi-Fu Chen.
\newblock Efficient pedestrian detection in top-view fisheye images using compositions of perspective view patches.
\newblock \emph{Image and Vision Computing}, 105:\penalty0 104069, 2021.

\bibitem[Cohen et~al.(2019)Cohen, Geiger, Koehler, and Welling]{cohenSphericalCNNs2019}
Taco~S. Cohen, Mario Geiger, Jonas Koehler, and Max Welling.
\newblock Spherical {{CNNs}}, 2019.

\bibitem[Fischer et~al.(2018)Fischer, Chang, and Demiris]{fischerRTGENERealTimeEye2018}
Tobias Fischer, Hyung~Jin Chang, and Yiannis Demiris.
\newblock {{RT-GENE}}: {{Real-Time Eye Gaze Estimation}} in {{Natural Environments}}.
\newblock In \emph{Computer {{Vision}} -- {{ECCV}} 2018}, pages 339--357. Springer International Publishing, Cham, 2018.

\bibitem[Funes~Mora et~al.(2014)Funes~Mora, Monay, and Odobez]{funesmoraEYEDIAPDatabaseDevelopment2014}
Kenneth~Alberto Funes~Mora, Florent Monay, and Jean-Marc Odobez.
\newblock {{EYEDIAP}}: A database for the development and evaluation of gaze estimation algorithms from {{RGB}} and {{RGB-D}} cameras.
\newblock In \emph{Proceedings of the {{Symposium}} on {{Eye Tracking Research}} and {{Applications}}}, pages 255--258, Safety Harbor Florida, 2014. ACM.

\bibitem[Ghosh et~al.(2024)Ghosh, Dhall, Hayat, Knibbe, and Ji]{ghoshAutomaticGazeAnalysis2024}
Shreya Ghosh, Abhinav Dhall, Munawar Hayat, Jarrod Knibbe, and Qiang Ji.
\newblock Automatic {{Gaze Analysis}}: {{A Survey}} of {{Deep Learning Based Approaches}}.
\newblock \emph{IEEE Transactions on Pattern Analysis and Machine Intelligence}, 46\penalty0 (1):\penalty0 61--84, 2024.

\bibitem[Harezlak and Kasprowski(2018)]{harezlakApplicationEyeTracking2018}
Katarzyna Harezlak and Pawel Kasprowski.
\newblock Application of eye tracking in medicine: {{A}} survey, research issues and challenges.
\newblock \emph{Computerized Medical Imaging and Graphics}, 65:\penalty0 176--190, 2018.

\bibitem[Kannala and Brandt(2004)]{kannalaGenericCameraCalibration2004}
J. Kannala and S. Brandt.
\newblock A generic camera calibration method for fish-eye lenses.
\newblock In \emph{Proceedings of the 17th {{International Conference}} on {{Pattern Recognition}}, 2004. {{ICPR}} 2004.}, pages 10--13 Vol.1, Cambridge, UK, 2004. IEEE.

\bibitem[Kasneci et~al.(2024)Kasneci, Gao, Ozdel, Maquiling, Thaqi, Lau, Rong, Kasneci, and Bozkir]{kasneciIntroductionEyeTracking2024}
Enkelejda Kasneci, Hong Gao, Suleyman Ozdel, Virmarie Maquiling, Enkeleda Thaqi, Carrie Lau, Yao Rong, Gjergji Kasneci, and Efe Bozkir.
\newblock Introduction to {{Eye Tracking}}: {{A Hands-On Tutorial}} for {{Students}} and {{Practitioners}}, 2024.

\bibitem[Ke et~al.(2024)Ke, Liu, Sokolikj, {Dahlstrom-Hakki}, and Israel]{keUsingEyetrackingEducation2024}
Fengfeng Ke, Ruohan Liu, Zlatko Sokolikj, Ibrahim {Dahlstrom-Hakki}, and Maya Israel.
\newblock Using eye-tracking in education: Review of empirical research and technology.
\newblock \emph{Educational technology research and development}, 72\penalty0 (3):\penalty0 1383--1418, 2024.

\bibitem[Kellnhofer et~al.(2019)Kellnhofer, Recasens, Stent, Matusik, and Torralba]{kellnhoferGaze360PhysicallyUnconstrained2019}
Petr Kellnhofer, Adria Recasens, Simon Stent, Wojciech Matusik, and Antonio Torralba.
\newblock Gaze360: {{Physically Unconstrained Gaze Estimation}} in the {{Wild}}.
\newblock In \emph{2019 {{IEEE}}/{{CVF International Conference}} on {{Computer Vision}} ({{ICCV}})}, pages 6911--6920, Seoul, Korea (South), 2019. IEEE.

\bibitem[Kumar et~al.(2023)Kumar, Eising, Witt, and Yogamani]{kumarSurroundviewFisheyeCamera2023}
Varun~Ravi Kumar, Ciaran Eising, Christian Witt, and Senthil Yogamani.
\newblock Surround-view {{Fisheye Camera Perception}} for {{Automated Driving}}: {{Overview}}, {{Survey}} and {{Challenges}}, 2023.

\bibitem[Li et~al.(2020)Li, Liu, Wang, Nishimura, and Kankanhalli]{liWeaklySupervisedMultiPersonAction2020}
Junnan Li, Jianquan Liu, Yongkang Wang, Shoji Nishimura, and Mohan~S. Kankanhalli.
\newblock Weakly-{{Supervised Multi-Person Action Recognition}} in 360{$^\circ$} {{Videos}}.
\newblock In \emph{2020 {{IEEE Winter Conference}} on {{Applications}} of {{Computer Vision}} ({{WACV}})}, pages 497--505, 2020.

\bibitem[Lu et~al.(2012)Lu, Sugano, Okabe, and Sato]{luHeadPosefreeAppearancebased2012}
Feng Lu, Yusuke Sugano, Takahiro Okabe, and Yoichi Sato.
\newblock Head pose-free appearance-based gaze sensing via eye image synthesis.
\newblock In \emph{Proceedings of the 21st {{International Conference}} on {{Pattern Recognition}} ({{ICPR2012}})}, pages 1008--1011, 2012.

\bibitem[Lu et~al.(2014)Lu, Okabe, Sugano, and Sato]{luLearningGazeBiases2014}
Feng Lu, Takahiro Okabe, Yusuke Sugano, and Yoichi Sato.
\newblock Learning gaze biases with head motion for head pose-free gaze estimation.
\newblock \emph{Image and Vision Computing}, 32\penalty0 (3):\penalty0 169--179, 2014.

\bibitem[Ogusu and Yamanaka(2019)]{ogusuLPMLearnablePooling2019}
Reo Ogusu and Takao Yamanaka.
\newblock {{LPM}}: {{Learnable Pooling Module}} for {{Efficient Full-Face Gaze Estimation}}.
\newblock In \emph{2019 14th {{IEEE International Conference}} on {{Automatic Face}} \& {{Gesture Recognition}} ({{FG}} 2019)}, pages 1--5, 2019.

\bibitem[Pfeuffer et~al.(2021)Pfeuffer, Alexander, and Gellersen]{pfeufferMultiuserGazebasedInteraction2021}
Ken Pfeuffer, Jason Alexander, and Hans Gellersen.
\newblock Multi-user {{Gaze-based Interaction Techniques}} on {{Collaborative Touchscreens}}.
\newblock In \emph{{{ACM Symposium}} on {{Eye Tracking Research}} and {{Applications}}}, pages 1--7, Virtual Event Germany, 2021. ACM.

\bibitem[Plopski et~al.(2023)Plopski, Hirzle, Norouzi, Qian, Bruder, and Langlotz]{plopskiEyeExtendedReality2023}
Alexander Plopski, Teresa Hirzle, Nahal Norouzi, Long Qian, Gerd Bruder, and Tobias Langlotz.
\newblock The {{Eye}} in {{Extended Reality}}: {{A Survey}} on {{Gaze Interaction}} and {{Eye Tracking}} in {{Head-worn Extended Reality}}.
\newblock \emph{ACM Computing Surveys}, 55\penalty0 (3):\penalty0 1--39, 2023.

\bibitem[Pulling et~al.(2024)Pulling, Tan, Hu, and Scherer]{pullingGeometryInformedDistanceCandidate2024}
Conner Pulling, Je~Hon Tan, Yaoyu Hu, and Sebastian Scherer.
\newblock Geometry-{{Informed Distance Candidate Selection}} for {{Adaptive Lightweight Omnidirectional Stereo Vision}} with {{Fisheye Images}}.
\newblock In \emph{2024 {{IEEE International Conference}} on {{Robotics}} and {{Automation}} ({{ICRA}})}, pages 12255--12261, 2024.

\bibitem[Qin et~al.(2025)Qin, Zhang, and Sugano]{qinUniGazeUniversalGaze2025}
Jiawei Qin, Xucong Zhang, and Yusuke Sugano.
\newblock {{UniGaze}}: {{Towards Universal Gaze Estimation}} via {{Large-scale Pre-Training}}, 2025.

\bibitem[Tan et~al.(2002)Tan, Kriegman, and Ahuja]{tanAppearancebasedEyeGaze2002}
Kar-Han Tan, D.J. Kriegman, and N. Ahuja.
\newblock Appearance-based eye gaze estimation.
\newblock In \emph{Sixth {{IEEE Workshop}} on {{Applications}} of {{Computer Vision}}, 2002. ({{WACV}} 2002). {{Proceedings}}.}, pages 191--195, 2002.

\bibitem[Wang et~al.(2019)Wang, Zhao, Su, and Ji]{wangGeneralizingEyeTracking2019}
Kang Wang, Rui Zhao, Hui Su, and Qiang Ji.
\newblock Generalizing {{Eye Tracking With Bayesian Adversarial Learning}}.
\newblock In \emph{2019 {{IEEE}}/{{CVF Conference}} on {{Computer Vision}} and {{Pattern Recognition}} ({{CVPR}})}, pages 11899--11908, 2019.

\bibitem[Wei et~al.(2023)Wei, Su, Wei, and Lu]{weiRotationalConvolutionRethinking2023}
Xuan Wei, Shixiang Su, Yun Wei, and Xiaobo Lu.
\newblock Rotational {{Convolution}}: {{Rethinking Convolution}} for {{Downside Fisheye Images}}.
\newblock \emph{IEEE Transactions on Image Processing}, 32:\penalty0 4355--4364, 2023.

\bibitem[Zhang et~al.(2022)Zhang, Liu, and Lu]{zhangGazeOnceRealTimeMultiPerson2022}
Mingfang Zhang, Yunfei Liu, and Feng Lu.
\newblock {{GazeOnce}}: {{Real-Time Multi-Person Gaze Estimation}}.
\newblock In \emph{2022 {{IEEE}}/{{CVF Conference}} on {{Computer Vision}} and {{Pattern Recognition}} ({{CVPR}})}, pages 4187--4196, New Orleans, LA, USA, 2022. IEEE.

\bibitem[Zhang et~al.(2015)Zhang, Sugano, Fritz, and Bulling]{zhangAppearanceBasedGazeEstimation2015}
Xucong Zhang, Yusuke Sugano, Mario Fritz, and Andreas Bulling.
\newblock Appearance-{{Based Gaze Estimation}} in the {{Wild}}.
\newblock In \emph{2015 {{IEEE Conference}} on {{Computer Vision}} and {{Pattern Recognition}} ({{CVPR}})}, pages 4511--4520, 2015.

\bibitem[Zhang et~al.(2017{\natexlab{a}})Zhang, Sugano, Fritz, and Bulling]{zhangItsWrittenAll2017}
Xucong Zhang, Yusuke Sugano, Mario Fritz, and Andreas Bulling.
\newblock It's {{Written All Over Your Face}}: {{Full-Face Appearance-Based Gaze Estimation}}.
\newblock In \emph{2017 {{IEEE Conference}} on {{Computer Vision}} and {{Pattern Recognition Workshops}} ({{CVPRW}})}, pages 2299--2308, 2017{\natexlab{a}}.

\bibitem[Zhang et~al.(2017{\natexlab{b}})Zhang, Sugano, Fritz, and Bulling]{zhangMPIIGazeRealWorldDataset2017}
Xucong Zhang, Yusuke Sugano, Mario Fritz, and Andreas Bulling.
\newblock {{MPIIGaze}}: {{Real-World Dataset}} and {{Deep Appearance-Based Gaze Estimation}}, 2017{\natexlab{b}}.

\bibitem[Zhang et~al.(2020)Zhang, Park, Beeler, Bradley, Tang, and Hilliges]{zhangETHXGazeLargeScale2020}
Xucong Zhang, Seonwook Park, Thabo Beeler, Derek Bradley, Siyu Tang, and Otmar Hilliges.
\newblock {{ETH-XGaze}}: {{A Large Scale Dataset}} for {{Gaze Estimation}} under {{Extreme Head Pose}} and {{Gaze Variation}}, 2020.

\bibitem[Zhu et~al.(2018)Zhu, Hu, Lin, and Dai]{zhuDeformableConvNetsV22018}
Xizhou Zhu, Han Hu, Stephen Lin, and Jifeng Dai.
\newblock Deformable {{ConvNets}} v2: {{More Deformable}}, {{Better Results}}, 2018.

\end{thebibliography}
}

\clearpage
\setcounter{page}{1}
\maketitlesupplementary

\section{MPSGaze360 Dataset Details}

This section provides additional information about the MPSGaze360 dataset. We supplement the main paper with the following details.

\subsection{MetaHuman Model Diversity}
The dataset includes 69 MetaHuman digital models. Statistics of their gender, age, skin tone, and ethnicity distributions are summarized in Table~\ref{tab:attr}.
All attribute annotations were automatically inferred by the GPT-4o model using front-view facial images. While factors such as illumination variation, visual ambiguity, and hallucination may introduce deviations from expert-defined labels, the aggregated statistics offer a reasonably faithful characterization of the overall distribution based on our observations.

\renewcommand{\arraystretch}{1.45}

\begin{table}[ht]
\centering
\caption{Demographic attribute statistics of the 69 MetaHuman characters used in the MPSGaze360 dataset.}
\label{tab:metahuman_stats}
\footnotesize

\begin{tabular}{l p{2.5cm} c c}
\hline
\textbf{Attribute} & \textbf{Category} & \textbf{Count} & \textbf{Percentage} \\
\hline

\multirow{2}{*}{Gender}
& Female & 32 & 46.38\% \\
& Male   & 37 & 53.62\% \\
\\[-12pt] \hline

\multirow{4}{*}{Age}
& 15--24 years        & 10 & 14.49\% \\
& 25--44 years        & 44 & 63.77\% \\
& 45--64 years        & 8  & 11.59\% \\
& 65 years and above  & 7  & 10.14\% \\
\\[-12pt] \hline

\multirow{5}{*}{Skin Tone}
& Fair          & 9  & 13.04\% \\
& Light Medium  & 26 & 37.68\% \\
& Medium        & 19 & 27.54\% \\
& Deep Medium   & 12 & 17.39\% \\
& Dark          & 3  & 4.348\% \\
\\[-12pt] \hline

\multirow{4}{*}{Ethnicity}
& African    & 23 & 33.33\% \\
& Caucasian  & 21 & 30.43\% \\
& East Asian & 16 & 23.19\% \\
& Other      & 9  & 13.04\% \\
\hline
\end{tabular}
\label{tab:attr}
\end{table}

\subsection{Diversity of Rendered Facial Appearance}
\begin{figure}[ht]
\centering
\includegraphics[width=\columnwidth]{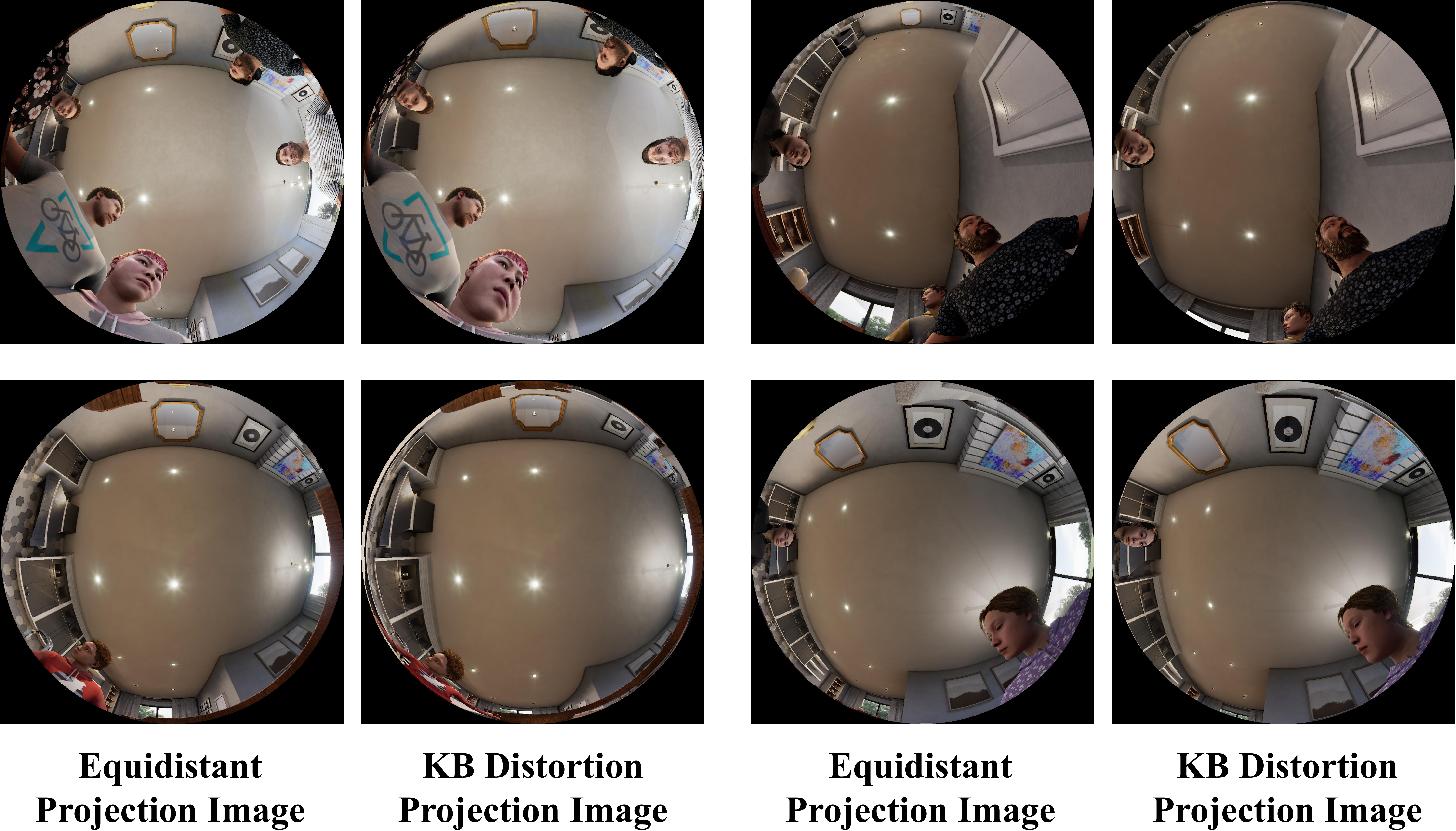}
\caption{
Comparison between images rendered using an equidistant projection and those rendered with a Kannala--Brandt (KB) distortion model.
}
\label{fig:distort}
\vspace{-3mm}
\end{figure}

Figure~\ref{fig:distort} presents close-up crops of faces from the dataset. The rendered images contain clear facial and ocular details, which are beneficial for learning causal gaze-related features.
Across samples, we observe substantial diversity in lighting conditions, identity appearance, head pose, gaze direction, subject–camera distance, and facial sharpness.

\begin{figure}[ht]
\centering
\includegraphics[width=\columnwidth]{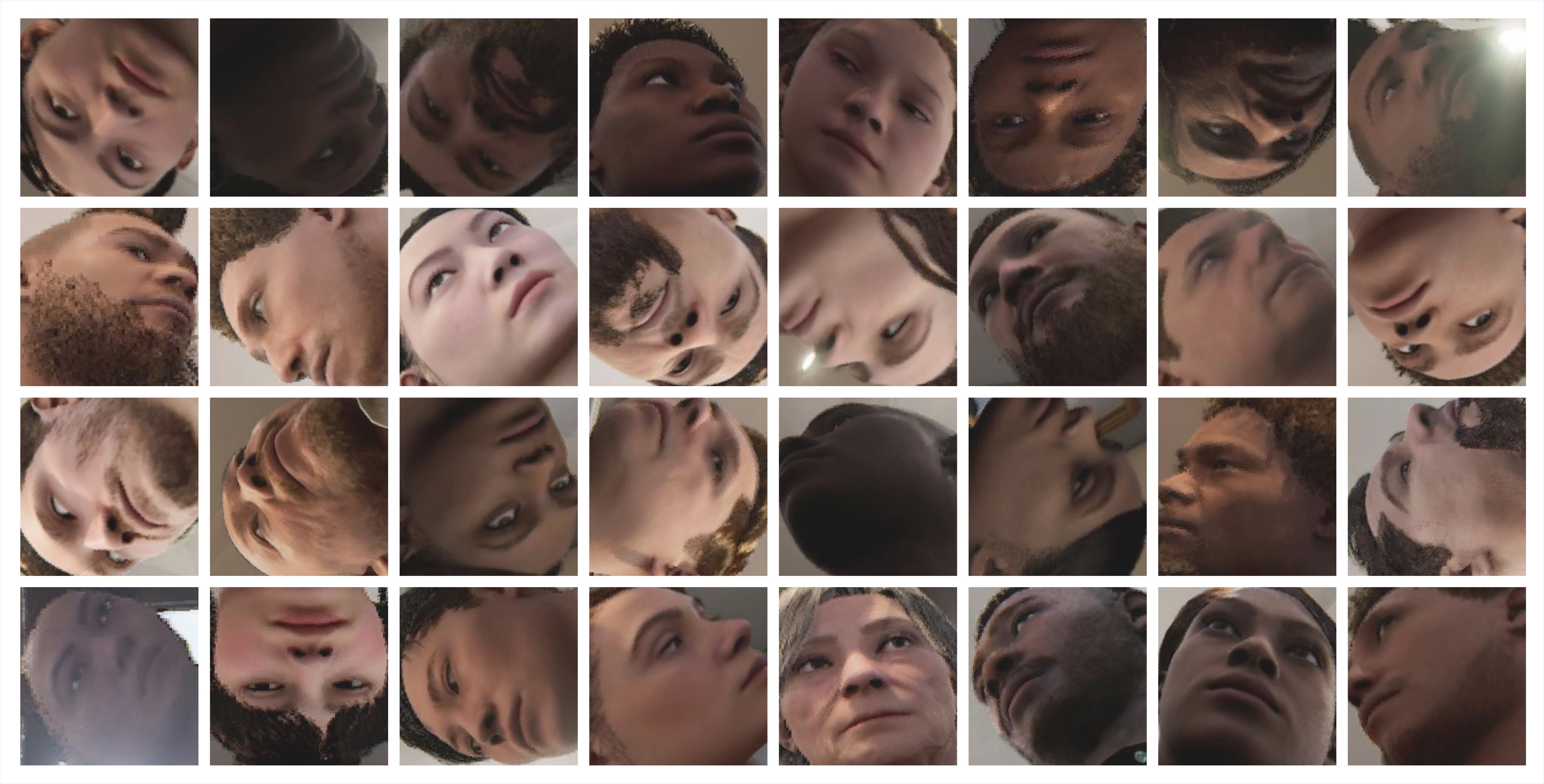}
\caption{
Examples of close-up facial crops from the MPSGaze360 dataset, illustrating the diversity in appearance, lighting, head pose, and gaze direction.
}
\label{fig:face_crop}
\vspace{-3mm}
\end{figure}

\subsection{Extended Fisheye Distortion Parameters}
Figure~\ref{fig:distort} further compares images rendered using an equidistant projection with those produced using a Kannala–Brandt distortion model with a larger distortion coefficient. In practice, many commercially available fisheye cameras exhibit relatively mild distortion, so models trained on equidistant fisheye images over a cropped 180° FOV are often sufficient for gaze estimation. However, for cameras with stronger distortion, our synthesis pipeline allows distortion parameters to be adjusted directly based on camera calibration, enabling the generation of new images and corresponding ground-truth annotations. Importantly, this process does not require regenerating the entire scene; a fast transformation can be applied using the stored 3D ground-truth. This flexibility allows the dataset to support rapid adaptation to diverse fisheye cameras in deployments.

\begin{table*}[tp]
\centering
\small
\caption{Impact of different hyperparameters and loss weights.} 
\label{tab:hyper}
\resizebox{1.8\columnwidth}{!}{
\begin{tabular}{lcccc}
\hline
Setting & Precision $\uparrow$ & Recall $\uparrow$ & Gaze Error ($^\circ$) $\downarrow$ & Adjusted Gaze Error ($^\circ$) $\downarrow$ \\
\hline
Baseline (lr=$10^{-3}$, glw=1.0) & 0.9992 & 0.9903 & 10.39 & 10.20 \\
lr=$5\times10^{-4}$ & 0.9975 &  0.9905 & 11.85 & 11.74 \\
lr=$2\times10^{-3}$  & 0.9781 & 0.9574 & 13.16 & 13.14 \\
glw=$0.5$  & 0.9877 &  0.9911 & 11.15 & 10.97 \\
glw=$5.0$  & 0.9928 & 0.9899 & 8.742 & 8.503 \\
glw=$10.0$  & 0.9989 & 0.9911 & 7.741 & 7.592 \\
\hline
\end{tabular}}
\end{table*}

\section{More Experiments}

\subsubsection{Robustness Across Face Resolutions}
\label{sec:resolution_robustness}

To evaluate robustness under varying face scales, we divide test samples into groups according to the detected face width and compute gaze errors for each group. As shown in Table~\ref{tab:face_width_errors}, low-resolution faces (30–60 px) lead to larger angular errors due to reduced detail.

\begin{table}[htbp]
\centering
\small
\setlength{\tabcolsep}{3pt}
\renewcommand{\arraystretch}{1.1}
\caption{Gaze error (°) under different face width intervals (pixels).}
\label{tab:face_width_errors}
\resizebox{\columnwidth}{!}{
\begin{tabular}{lccccc}
\hline
\textbf{Setting} & \textbf{30--60} & \textbf{60--90} & \textbf{90--120} & \textbf{120--150} & \textbf{$>$150} \\
\hline
w/o RotConv, w/o ldmks     & 13.91 & 11.65 & 12.02 & 11.75 & 11.70 \\
RotConv only            & 12.50 & 10.73 & 11.09 & 10.66 & 12.17 \\
RotConv + face ldmks     & 11.27 & 9.230 & 9.957 & 9.323 & 10.59 \\
RotConv + eye ldmks      & 11.03 & 8.293 & 8.617 & 9.368 & 7.569 \\
RotConv + face + eye ldmks & 10.80 & 8.431 & 8.811 & 8.919 & 8.052 \\
\hline
\end{tabular}}
\end{table}

\subsubsection{Metrics w.r.t Head Pose and Distance}

Tables~\ref{tab:head_pose} and~\ref{tab:head_distance} report the gaze estimation errors under different head pose yaw angles and head distances, respectively. Specifically, Table~\ref{tab:head_pose} shows the gaze error across absolute yaw intervals from $0^\circ$ to $90^\circ$, while Table~\ref{tab:head_distance} presents errors for varying head distances in centimeters. 

The results indicate that gaze error increases moderately with larger head rotations and greater distances, reflecting the inherent difficulty of predicting gaze under extreme poses or when faces occupy fewer pixels.

\begin{table}[htbp]
\centering
\small
\setlength{\tabcolsep}{3pt}
\renewcommand{\arraystretch}{1.1}
\caption{Gaze error (°) under different absolute head yaw angles.}
\label{tab:head_pose}
\resizebox{\columnwidth}{!}{
\begin{tabular}{lcccccc}
\hline
\textbf{Head pose yaw} & \textbf{0--10} & \textbf{10--20} & \textbf{20--30} & \textbf{30--45} & \textbf{45--60} & \textbf{60--90} \\
\hline
     & 9.99 & 10.23 & 10.68 & 10.13 & 10.54 & 11.07 \\

\hline
\end{tabular}}
\end{table}

\begin{table}[htbp]
\centering
\small
\setlength{\tabcolsep}{3pt}
\renewcommand{\arraystretch}{1.1}
\caption{Gaze error (°) under different head distances (cm).}
\label{tab:head_distance}
\resizebox{0.7\columnwidth}{!}{
\begin{tabular}{lcccc}
\hline
\textbf{Head distance} & \textbf{30--50} & \textbf{50--70} & \textbf{70--90} & \textbf{$>$90} \\
\hline
     & 9.506 & 9.875 & 10.47 & 12.73 \\

\hline
\end{tabular}}
\end{table}

\subsection{Hyperparameters and Loss Weights}

In the main experiments, we use an initial learning rate of $10^{-3}$ and set all loss weights to 1 to avoid any task-specific weighting heuristics. In the supplementary material, we additionally retrain the model with different learning rates and gaze loss weights (glw), with results reported in Table~\ref{tab:hyper}. We observe that adjusting the learning rate generally has a negative impact on performance, whereas increasing the gaze loss weight can further improve the model's accuracy.

\section{Additional Discussions}

The additional experiments presented in this supplementary material further validate the robustness and generalization of GazeOnce360 under various settings, including different hyperparameters, loss weights, head poses, and distances. These results provide a deeper understanding of the model's behavior and sensitivity, complementing the main paper.

We also observe that the fisheye image resolution limits gaze estimation for subjects located more than 2 meters from the camera, which in turn constrains the practical deployment of the system. Moreover, accurate estimation requires that subjects face the camera to some extent; extreme gaze angles relative to the camera, as well as occlusions between individuals, can lead to unreliable or missing gaze predictions. These factors represent inherent limitations of the current system.

Future work could explore real-world deployment considerations, domain adaptation techniques, fisheye gaze target estimation, and ethical safeguards.

\section{Broader Impact}

From a broader impact perspective, 360° multi-person gaze estimation has potential applications in human-computer interaction, social behavior analysis, and collaborative robotics. At the same time, as with all vision-based human analysis systems, care must be taken to respect privacy and prevent misuse in sensitive contexts.

\end{document}